\documentclass[preprint,12pt]{elsarticle}

\usepackage[utf8]{inputenc}
\usepackage[T1]{fontenc}
\usepackage{subcaption}
\usepackage{booktabs}
\usepackage{multirow}
\usepackage{url}
\usepackage{tikz}
\usepackage{paralist}
\usepackage{ifthen}
\usepackage{amsmath}
\newcommand{\CC}[1][]{$\text{C\hspace{-.25ex}}^{_{_{_{++}}}}
\ifthenelse{\equal{#1}{}}{}{\text{\hspace{-.625ex}#1}}$}

\usepackage{bm}
\usepackage{hyperref}
\usepackage{amsmath}
\usepackage{amssymb}
\usepackage{amsthm}
\usepackage{amsfonts}
\usepackage{thmtools}		
\usepackage{mleftright}
\usepackage{stmaryrd}
\usepackage{soul}
\usepackage{nicefrac}
\usepackage{fancyref}
\usepackage{algorithm}
\usepackage{algorithmicx}
\usepackage[noend]{algpseudocode}

\let\originalleft\left
\let\originalright\right
\renewcommand{\left}{\mathopen{}\mathclose\bgroup\originalleft}
\renewcommand{\right}{\aftergroup\egroup\originalright}

\theoremstyle{definition}

\usepackage{thm-restate}
\usepackage[mathic=true]{mathtools}
\usepackage{fixmath}
\usepackage{siunitx}

\usepackage{pifont}

\usepackage{enumitem}
\setlist[enumerate]{itemsep=0.2ex, topsep=0.5\topsep}
\setlist[description]{itemsep=0.2ex, topsep=0.5\topsep}
\setlist[itemize]{itemsep=0.2ex, topsep=0.5\topsep}

\usepackage[capitalise,noabbrev]{cleveref}   

\usepackage{microtype}
\usepackage{ellipsis}

\usepackage[scaled=0.86]{helvet}
\usepackage{lmodern}

\usepackage{soul}

\newcommand{\old}[1]{}
\newcommand{\new}[1]{{\color{black}{#1}}}

\journal{Neural Networks}

\newcommand{\bi}{\mathbf{i}}

\newcommand{\bw}{\mathbf{w}}
\newcommand{\bx}{\mathbf{x}}

\newcommand{\bD}{\mathbf{D}}

\newcommand{\bF}{\mathbf{F}}

\newcommand{\bW}{\mathbf{W}}

\newcommand{\bxzero}{\mathbf{x}^{(0)}}
\newcommand{\bxd}{\mathbf{x}^{(d)}}
\newcommand{\bzd}{\mathbf{z}^{(d)}}

\newcommand{\bxdmone}{\mathbf{x}^{(d-1)}}
\newcommand{\bbd}{\mathbf{b}^{(d)}}
\newcommand{\bWd}{\mathbf{W}^{(d)}}
\newcommand{\Wijd}{W_{ij}^{(d)}}
\newcommand{\bid}{b_i^{(d)}}
\newcommand{\bhd}{\mathbf{h}^{(d)}}
\newcommand{\hijd}{\mathbf{h}^{(d)}}

\newcommand{\diag}{\operatorname{diag}}
\newcommand{\tildebhd}{\tilde{\mathbf{h}}^{(d)}}
\newcommand{\vard}{\nu^{(d)}}
\newcommand{\covd}{c_{12}^{(d)}}
\newcommand{\vardpone}{\nu^{(d+1)}}
\newcommand{\covdpone}{c_{12}^{(d+1)}}
\newcommand{\varstar}{\nu^*}

\begin{document}

\begin{frontmatter}



\title{Revisiting Deep Information Propagation: \\
Fractal Frontier and Finite-size Effects}



\author[1]{Giuseppe Alessio D'Inverno\corref{cor1}}


\author[2]{Zhiyuan Hu}

\author[3]{Leo Davy}

\author[2]{Michael Unser}

\author[1]{Gianluigi Rozza}

\author[2]{Jonathan Dong}

\affiliation[1]{organization={MathLab, International School for Advanced Studies (SISSA)},
           addressline={Via Bonomea 265}, 
           city={Trieste},
           postcode={34136}, 
           country={Italy}}
           
\affiliation[2]{organization={Biomedical Imaging Group, École Polytechnique Fédérale de Lausanne},
            addressline={Station 17}, 
            city={Lausanne},
            postcode={1015}, 
            country={Switzerland}}           
\affiliation[3]{organization={Laboratoire
de Physique de l’ENS Lyon, CNRS UMR 5672},
           addressline={46 All. d'Italie}, 
           city={Lyon},
           postcode={69007}, 
           country={France}}
\cortext[cor1]{Corresponding author. Email address: gdinvern@sissa.it}

\begin{abstract}
Information propagation characterizes how input correlations evolve across layers in deep neural networks.  This framework has been well studied using mean-field theory, which assumes infinitely wide networks. However, these assumptions break down for practical, finite-size networks. In this work, we study information propagation in randomly initialized neural networks with finite width and reveal that the boundary between ordered and chaotic regimes exhibits a fractal structure. This shows the fundamental complexity of neural network dynamics, in a setting that is independent of input data and optimization. To extend this analysis beyond multilayer perceptrons, we leverage recently introduced Fourier-based structured transforms, 
and show that information propagation in convolutional neural networks also follow the same behavior. 
\old{Our}\new{In practice, our} 
investigation highlights the importance of finite network depth with respect to the tradeoff between separation and robustness. 
\new{We also show that fractal patterns are observed for information propagation in the backward pass, i.e., backpropagation from the last to the first layer of finite-size networks.}
\end{abstract}


\begin{highlights}
\item The boundary between ordered and chaotic information propagation regimes in finite-width neural networks exhibits a fractal structure, highlighting fundamental complexity independent of input data \old{or optimization}.
\item Information propagation in convolutional neural networks follows the same fractal behavior observed in multilayer perceptrons, demonstrated through the use of Fourier-based structured transforms.
\item The study emphasizes the critical role of finite network depth in balancing the trade-off between separation and robustness in deep neural networks.
\end{highlights}


\end{frontmatter}

\section{Introduction}
The remarkable success of deep neural networks across a wide range of machine learning applications is well established. While they have demonstrated an impressive ability to learn complex representations of images, text, and other modalities, our theoretical understanding of these mathematical objects remains limited. This difficulty stems from the intricate interplay between linear layers and non-linear activations, which makes it challenging to analyze these models using conventional theoretical tools. As a consequence, the design of neural network architectures often remains more of an art than a science. Nonetheless, having clear and well-founded metrics is crucial \old{---for guiding} \new{to guide} the choice of hyperparameters, {\old{comparing} \new{compare}} architectures, and \old{informing} \new{inform} the design of new models. 

Information propagation \cite{schoenholz2017deep} characterizes how differences in network inputs evolve as they pass through the layers of a neural network. In the case of randomly initialized networks, a mean-field analysis reveals an \textit{order-to-chaos} transition when varying the scale of the network weights: in the \textit{ordered} phase, input signals decay exponentially with depth, while in the \textit{chaotic} phase, information can successfully propagate through the network. These two regimes are closely linked to network expressivity and backpropagation \cite{poole2016exponential}. Notably, the critical frontier between order and chaos has emerged as a particularly promising operating regime. It can be derived explicitly by the study of fixed points of the mean-field asymptotic equations. Parallel works have introduced similar mean-field approaches to the recurrent setting \cite{dong2020reservoir}, where information propagation is replaced by the notion of stability of the underlying dynamical system \cite{dong2022asymptotic}. While all previous studies are typically conducted in the \old{asymptotic} \new{mean-field} limit of infinitely wide, randomly connected networks, the framework can also be adapted to more general network topologies \cite{d2025comparison}. 

Interestingly, recent studies have reported that in the finite-width regime, the boundary of a similar order-to-chaos phase transition for training dynamics exhibits \textit{fractal} complexity \cite{sohl2024boundary, torkamandi2025mapping}. Fractals are complex geometric patterns that emerge from iterative processes, characterized by self-similarity across scales. Such structures have been observed along the transition between stable and divergent training dynamics as learning rates are varied, both in single hidden-layer networks \cite{sohl2024boundary} and in transformer architectures \cite{torkamandi2025mapping}. In both cases, the reported fractal dimensions lie strictly between 1 and 2, indicating a nontrivial geometric structure. This fractal behavior highlights the intricate nature of neural network training, emphasizing the non-monotonicity and irregularity of the hyperparameter landscape. 

In this work, we report and investigate the emergence of fractal structure for deep information propagation in finite-size neural networks. 
Our contributions are three-fold:
\begin{itemize}
\item First, we extend the class of analyzable network topologies beyond multilayer perceptrons (MLPs), including convolutional neural networks (CNNs) and recently proposed Fourier-based structured transforms \cite{hu2025structured}---efficient linear layers composed of Fourier transforms and element-wise multiplications---which interpolate between fully connected random layers \cite{yu2016orthogonal} and random convolutions \cite{hu2025structured}. We show that the mean-field approximation is valid for all these architectures. We systematically study deep information propagation by varying the variances of internal weights and biases. 
\item Second, we characterize the boundary between ordered and chaotic regimes in the finite-size setting and quantify its fractal dimension. Our findings suggest a much richer and more intricate dynamical behavior than what mean-field theory predicts, revealing a highly sensitive dependence on hyperparameters. Unlike previous fractal structures highlighted in neural network training dynamics, our approach relies on a fundamental metric that is independent of datasets and optimization algorithms, reflecting intrinsic properties of the network architecture itself.
\item \old{Finally, we} \new{We} discuss the tradeoff between information propagation and noise robustness---a challenge for very deep architectures that motivates the use of finite-depth networks.
\item \new{Finally, we also introduce the concept of information propagation during backpropagation and show that it exhibits a similar fractal boundary between the ordered and chaotic regimes.}
\end{itemize}

\section{Background and definitions}

\subsection{Neural network topologies}

Among neural network architectures, multilayer perceptrons serve as the canonical example of fully connected feedforward models.
An MLP consists of $D$ fully-connected layers of size $N$, each followed by an element-wise nonlinear activation function. 
The input is denoted as $\bxzero = \bi \in \mathbb{R}^N$ and for each layer $d = 1, \ldots, D$ the pre-activations $\bzd \in \mathbb{R}^N$ and activations $\bxd \in \mathbb{R}^N$ are given by:
\begin{equation}
    \label{eq: mlp definition}
    \bzd = \sigma_w \bWd \bxdmone + \sigma_b \bbd, \qquad \bxd = f(\bzd),
\end{equation}
where $\bWd \in \mathbb{R}^{N\times N}$ and $\bbd \in \mathbb{R}^N$ denote the weight matrix and bias of the $d$-th layer, respectively, and $f$ is an element-wise activation function. 
We chose for conciseness an input the size of the network, but it can be generalized to arbitrary inputs by modifying the dimensions of $\bW^{(1)}$. 
The random setting corresponds to the case where all weights and biases are i.i.d. Gaussian random variables $\Wijd \sim \mathcal{N}(0, 1 / N)$ and $\bid \sim \mathcal{N}(0, 1)$ for all layers $d\geq1$. 

Eq. \eqref{eq: mlp definition} also describes the update equation of a Recurrent Neural Network (RNN). There, $d$ denotes the time index and $\bxd$ the network state at time $d$, while $\bW$ are the (time-independent) internal network weights and $\bbd$ the input at time $d$. Since RNNs are notoriously hard to train, Reservoir Computing fixes all the internal weights $\bW$ at random, only training a linear output layer \cite{jaeger2001echo}. This class of algorithm has historically found many applications \cite{lukovsevivcius2012reservoir} and is now a promising avenue for large-scale efficient physical implementations \cite{tanaka2019recent, rafayelyan2020large}. Unrolling through time, this corresponds to a random MLP with a fixed weight matrix shared between layers. 

The form of Eq. \eqref{eq: mlp definition} as a succession of linear layers and nonlinear activations also include convolutional neural networks. For CNNs, the linear operator $\bWd$ represents a convolution defined as:
\begin{equation}
    \label{eq: convolution definition}
    \bWd \bxdmone = \bhd * \bxdmone
\end{equation}
where $*$ denotes the convolution operator and $\bhd \in \mathbb{R}^K$.\footnote{We focus in this study on 1D convolutions but conclusions extend to the 2D case.}
Compared to the fully-connected case, this linear operator possesses a Toeplitz-like structure with correlated entries.
A random convolution can be represented as $\hijd \sim \mathcal{N}(0, 1 / K)$. 
Convolutions can be computed in real space, which is more efficient for small kernel sizes. Alternatively, they can be implemented in Fourier space:
\begin{equation}
    \label{eq: Fourier convolution}
    \bhd * \bxdmone = \bF \diag\left(\tildebhd\right) \bF^H \bxdmone
\end{equation}
with $\bF$ the Discrete Fourier Transform operator, $H$ the Hermitian conjugate, and $\tildebhd$ the Fourier transform of the kernel $\bhd$.
As the FFTs can be computed in $\mathcal{O}(N \log N)$ complexity and the diagonal operator is an elementwise multiplication, this can speed up computations for \old{larger}\new{large} kernel sizes. 
It also modifies the boundary conditions from zero-padding to periodic boundary conditions. 

Interestingly, structured random models make the link between random MLPs and CNNs. 
They are defined as follows:
\begin{equation}
    \label{eq: structured random model}
    \bWd = \mathrm{Re}\left(\prod_{i=1}^N (\bF \bD_i^{(d)}) [\bF]\right)
\end{equation}
with $\mathrm{Re}$ the real part operator and $\bD_i^{(d)}$ random diagonal matrices for $i = 1, \ldots, N$, where $[F]$ denotes an optional Fourier Transform. As seen in Eq. \eqref{eq: Fourier convolution}, the case $N=1$ with the optional Fourier transform is similar to a random convolution. In \cite{hu2025structured}, it was observed that for $N=2$, such transforms emulate fully-connected i.i.d. random matrices for their particular phase retrieval equation. Additional examples of structured transforms emulating i.i.d. random transforms include random features \cite{yu2016orthogonal} and Reservoir Computing \cite{dong2020reservoir}. Other variants train parameters in these structured transforms to speed up dense matrix computation \cite{moczulski2015acdc,dao2019learning}. These previous studies employed Hadamard or cosine transforms which are similar in terms of complexity, but we choose \old{the Fourier one as it presents} \new{Fourier transforms as they enable} the analogy with convolutions for $N=1$. 

All these network architectures are introduced as they present the full spectrum of network topologies from MLP to CNN with small kernel size. We want to investigate how they behave in terms of information propagation, in the deep case when the number of layers $D$ is large. 

\subsection{Deep information propagation}

\begin{figure}[t!]
    \centering
    \includegraphics[width=0.9\linewidth]{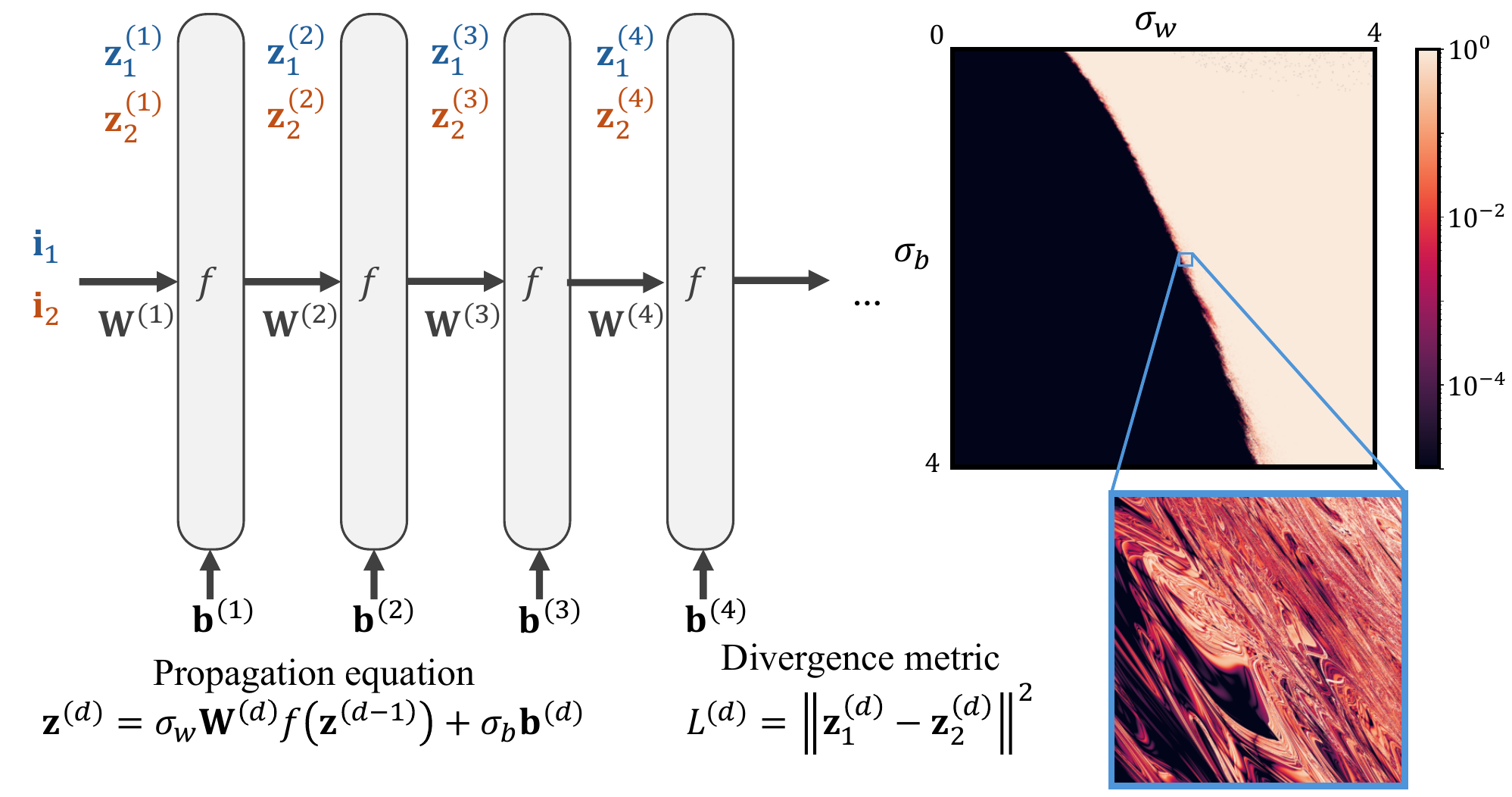}
    \caption{Graphical abstract. (Left) Information propagation experiment. Two different inputs are fed in a feedforward network and the divergence metric is defined as the distance between pre-activations. (Right) Fractal behavior of the information propagation landscape. The divergence metric \new{$L^{(D)}$} as a function of $\sigma_w$ and $\sigma_b$ (at depth $D = 10^3$ for a CNN of size $N = 100$ with $\operatorname{erf}$ activation function) exhibits a fractal structure for the boundary between stability and information propagation. }
    \label{fig:abstract}
\end{figure}

Information propagation through the network can be interpreted geometrically as the separation of different input signal trajectories as they propagate through the network layers, as depicted in Fig. \ref{fig:abstract}. Given a pair of input signals $\bi_1, \bi_2 \in \mathbb{R}^N$, the network produces corresponding activations $\bxd_1$ and $\bxd_2$ at each layer $d = 1, \ldots, D$. The Euclidean distance between the layer pre-activations quantifies the divergence of these trajectories:
\begin{equation}
\label{eq: propagation metric}
L^{(d)} = \left\|\bzd_1 - \bzd_2\right\|^2.
\end{equation}
We say that information propagates through the network if the output responses remain distinguishable, i.e. if $L^{(D)} > \tau$ for some constant threshold $\tau > 0$. 

This was studied previously in \cite{schoenholz2017deep} for the case of random MLPs, deriving mean-field equations in the limit of infinitely-wide networks. These iterated 1D equations yield simple fixed points iterations, the study of whom determines whether information propagates or vanishes in deep networks. 
The principle is the following: we rewrite Eq. \eqref{eq: propagation metric} as $L^{(d)} = \left\|\bzd_1\right\|^2 + \left\|\bzd_2\right\|^2 - 2 \left(\bzd_1\right)^H \bzd_2$ and the mean-field approximation replaces these stochastic quantities by their expectation values \cite{poole2016exponential}, 
\begin{equation}
    \label{eq: mean field divergence}
    L^{(d)} \approx \mathcal{L}^{(d)} = 2 \left(\vard - \covd\right),
\end{equation}
with the variance $\vard = \operatorname{Var}(\bzd_1) = \operatorname{Var}(\bzd_2)$ and covariance $\covd = \operatorname{Cov}(\bzd_1, \bzd_2)$, that are updated according to:
\begin{align}
    \label{eq: update variance d}
    \vardpone &= \sigma_w^2 \int f^2\left(\sqrt{\vard} z\right) p(z) dz + \sigma_b^2 \\
    \covdpone &= \sigma_w^2 \int f\left(\sqrt{\vard} z_1\right) f\left(\covd z_1 + \sqrt{\vard - \covd} z_2\right) p(z_1) p(z_2) dz_1 dz_2 + \sigma_b^2,
\end{align}
for scalar normal random variables $z$, $z_1$, $z_2$ following the distribution $p(u) = \frac{1}{\sqrt{2 \pi}} e^{-u^2 / 2}$. 
Here the previous preactivations of layer $d$ have been replaced by scalar gaussian random variables.

We thus introduce the functions $g$ and $h$ such that $\vardpone = g(\vard)$ and $\covdpone = h(\vard, \covd)$. 
The variance $\vard$ converges to the fixed point $\varstar$ of $g$ and $\covd$ converges to the smallest fixed point of $h(\varstar, \cdot)$. If the two are equal, $\mathcal{L}^{(d)}$ converges to zero, otherwise it converges to a non-zero value. 
We obtain a stable and a chaotic phase: in the stable phase, information does not propagate, while it does in the chaotic one. 
The boundary between information propagation and contractant activations is then defined for a fixed threshold $\tau \geq 0$ as:
\begin{equation}\label{eq:S_tau}
    S_\tau = \{(\sigma_r, \sigma_b) : \mathcal{L}(\sigma_r, \sigma_b) = \tau\}.
\end{equation}

The mean-field approximation in Eq. \eqref{eq: mean field divergence} corresponds to the limit of infinitely-wide networks.  
Similar equations have been derived in \cite{dong2020reservoir, dong2022asymptotic, d2025comparison}. Explicit convergence rates to the mean-field approximations have been derived in \cite{dong2020reservoir}, assuming bounded activations and contractant dynamics. 
Interestingly, convergence to the mean-field equations is not assured theoretically, one needs to add an additional contractive assumption (i.e. small $\sigma_w$) to ensure that the variance of the iterated random maps asymptotically tends to zero. In practice, we observe convergence over a wider range of settings even for large network weights, as long as the activation function is bounded.
Interestingly, this was developed for the recurrent case of Reservoir Computing. There, stable dynamics are preferred: after a fade-in period, the network state should not depend on an arbitrary initialization. Therefore, stability is desired, which corresponds to $L = 0$, the opposite of information propagation in feedforward architectures. 

\new{\subsection{Information propagation for backpropagation}

The concept of information propagation can also be extended to backpropagation, to investigate whether information about the loss function is back propagated through the whole network. 
The experiment is designed as follows: a single random input vector $\bi$ is fed in the previous neural network, initialized with random weights $\bWd$ and biases $\bbd$. 
The activations of the last layer $\bx^{(D)}$ are used to compute two linear loss functions $\ell_1 = \left(\bw^\text{out}_1\right)^H \bx^{(D)}$ and $\ell_2 = \left(\bw^\text{out}_2\right)^H \bx^{(D)}$ for two independent Gaussian random vectors $\bw^\text{out}_1$ and $\bw^\text{out}_2$. 
The gradients with respect to the last layer activations are
\begin{equation}
    \frac{\partial \ell_1}{\partial \bx^{(D)}} = \bw^\text{out}_1 
    \qquad \text{and}
    \qquad \frac{\partial \ell_2}{\partial \bx^{(D)}} = \bw^\text{out}_2.
\end{equation}
We then backpropagate both gradients through the network and compute the L2 distance between the gradients with respect to the first layer activations:
\begin{equation}
L'= \left\|\frac{\partial \ell_1}{\partial \bx^{(0)}} - \frac{\partial \ell_1}{\partial \bx^{(0)}}\right\|^2.
\end{equation}

This metric is computed for different values of the variances of internal weights and biases. When $L'$ converges to zero as the network depth increases, the gradients of the first layers do not depend on the loss function. In this case, information is not backpropagated through the network; such a property is undesirable as the gradients of the first layers are not informative. 

Compared to \cite{sohl2024boundary}, our study focuses on the initial backpropagation step of randomly-initialized networks, and does not depend on a particular iterative optimization algorithm. 
Up to our knowledge, there is no mean-field theory of this particular information backpropagation experiment, which could be an interesting subject for future studies. 
}

\subsection{Fractal dimension}

To study the fractal behavior of the information propagation frontier, let us start by defining metrics to characterize it. 
Fractals are intricate geometric shapes that display fine structures at arbitrarily small scales \cite{falconer_fractal_2006, strogatz2018nonlinear, abry2009scaling}. Some fractals exhibit self-similarity, as seen in the Sierpinski triangle or the von Koch curve, making it straightforward to compute their \textit{fractal dimension}---a measure of the fractal's complexity. However, for more complex and non-self-similar geometries, the fractal dimension must be estimated using more general measurement tools. 

\begin{figure}[t!]
    \centering
    \includegraphics[width=\linewidth]{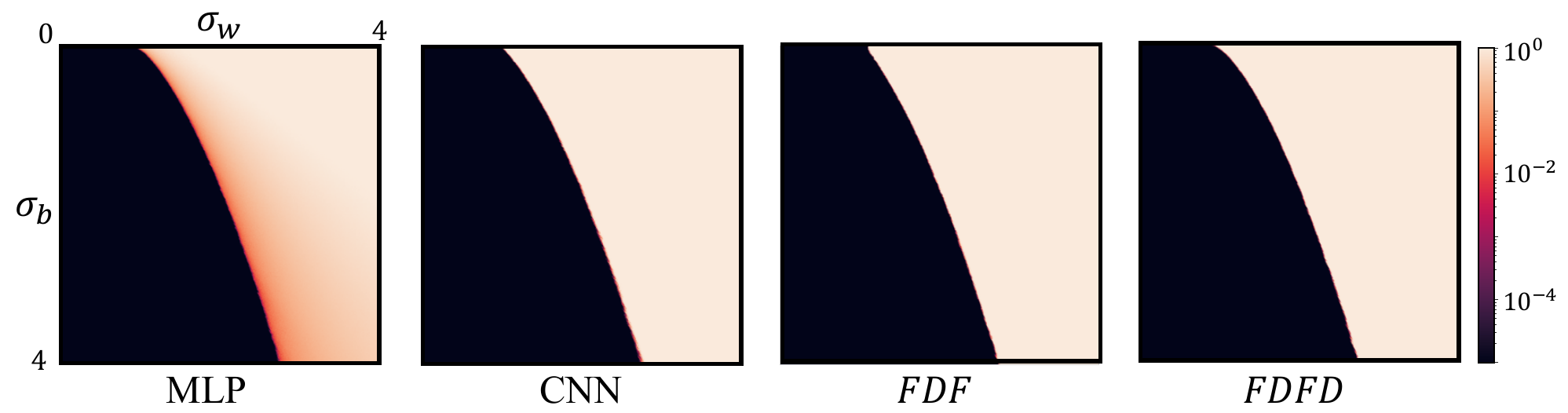}
    \caption{Universality of the frontier between stability and information propagation. Divergence metric \new{$L^{(D)}$} as a function of $\sigma_w$ and $\sigma_b$ for random fully-connected (MLP), random convolutional (CNN), or structured random of the form $W = FDF$ or $W = FDFD$ with Fourier transforms and random diagonal matrices. This has been computed for depth $D=10^3$ and width $N=10^3$.}
    \label{fig: asymptotic 2d}
\end{figure}

A fine-grained estimate of the fractal dimension of a geometric shape $S$ is given by the \textit{Hausdorff dimension} (see e.g. \cite[Chapter 2]{falconer_fractal_2006} or \cite[Chapter 1]{abry2009scaling}), that intuitively quantifies the rate at which the number of balls of radius at most $\delta$ required to cover $S$ increases as $\delta$ decreases. Nevertheless, this estimate can be simplified into the \textit{Minkowski dimension}, also called \textit{box dimension} \cite[Chapter 2]{falconer_fractal_2006}. We consider  $N(\varepsilon)$ the minimum number of cubes of side $\varepsilon$ required to cover $S$, and the box dimension $\Delta$ is the exponent such that $N(\varepsilon) \approx \varepsilon^{-\Delta}$, or more rigorously:
\begin{equation}
    \Delta(S) = \limsup_{\varepsilon\to 0} \frac{\ln N(\varepsilon)}{|\ln \varepsilon|}.
\end{equation}

The box dimension has been used in previous work to observe and quantify fractal structures along the transition between stable and divergent training dynamics when different learning rates are used either for the weight update in single hidden-layer networks \cite{sohl2024boundary} or for the attention layer and the other parameter weights in transformer architectures \cite{torkamandi2025mapping}. Our work takes inspiration from these useful insights, focusing instead on the variance of weights and biases rather than on the learning rates used in the training phase, therefore offering a perspective that is independent from the samples fed to the model or the optimization algorithm adopted.

\section{Results}\label{sec:results}


\subsection{Asymptotic deep information propagation for convolutional topologies}\label{sec: asymptotic_info_prop}

\begin{figure}[t]
    \centering
    \includegraphics[width=\linewidth]{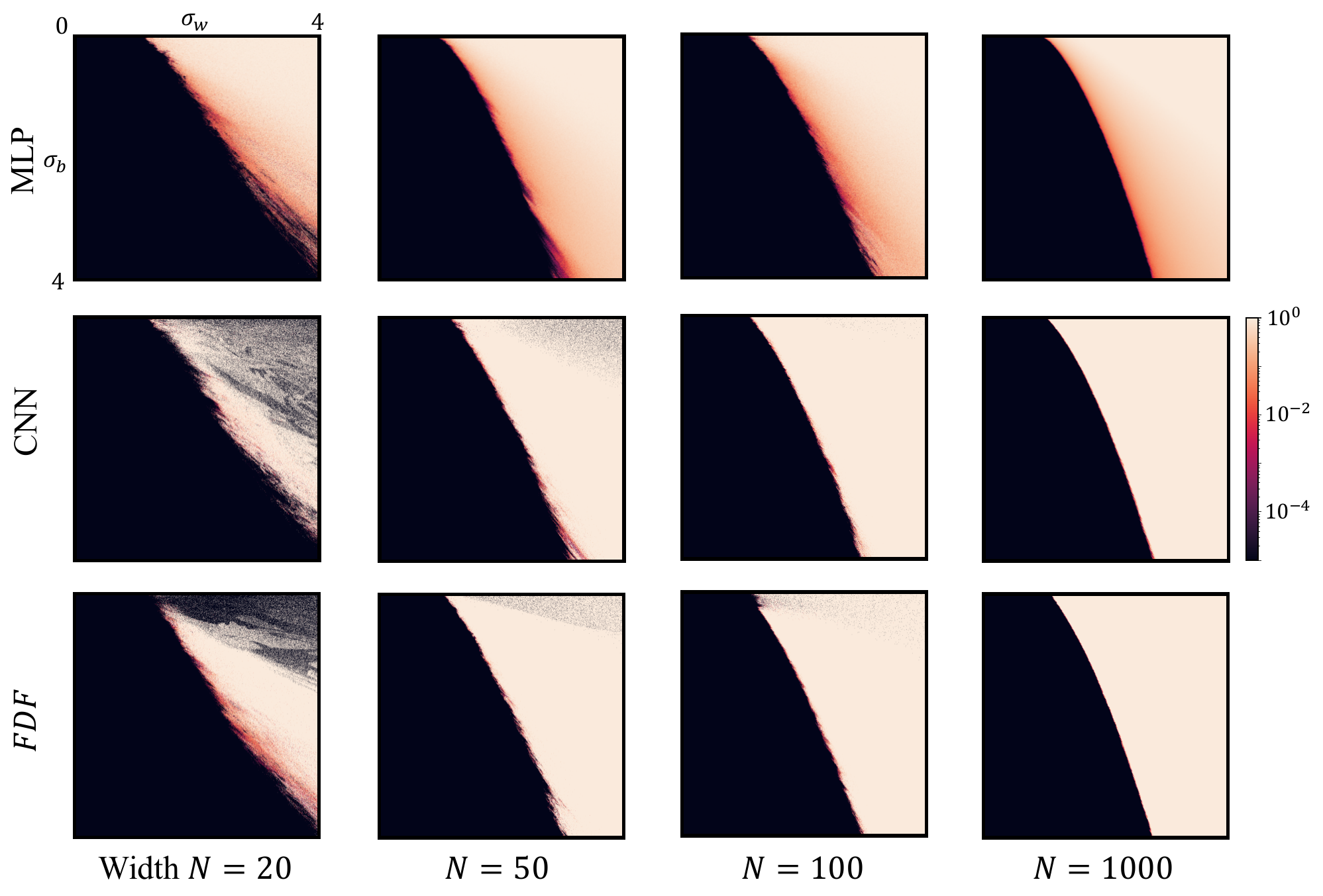}
    \caption{Convergence towards the mean-field limit of the different architectures. Examples of information propagation landscape for different widths, from small ($N=20$) to large ($N = 10^3$), and different architectures (MLP, CNN, structured random).}
    \label{fig:different-widths}
\end{figure}

We first investigate how deep information propagation converges to the same asymptotic limit across different network architectures. To do so, we generate two independent unit-norm Gaussian input vectors $\bxd_1, \bxd_2 \in \mathbb{R}^N$, and propagate them through randomly-initialized networks as defined in Eq. \eqref{eq: mlp definition}. The divergence metric $L^{(D)}$ is tracked while varying the internal weights variance $\sigma_w$ and bias variance $\sigma_b$. We set $N=10^3$, a large-enough value to closely approximate the infinite-width regime described by mean-field theory, and fixed the network depth at $D = 10^3$, averaging over the last 20 layers. The activation function used was $f = \operatorname{erf}$, a sigmoid non-linearity that enables analytical treatment in mean-field frameworks. 

Each $10^3 \times 10^3$ image in Fig. \ref{fig: asymptotic 2d} required about one hour on an NVIDIA V100 GPU, parallelizing over rows of the image. For convolutional architectures, the kernel size matched the network width and circular boundary conditions were applied. To show the universality of the mean-field limit, structured random transforms were also evaluated: $FDF$ corresponds to the case $N=1$, including a final Fourier transform, and $FDFD$ to $N=2$, excluding the final Fourier transform. In all structured cases, diagonal matrix elements were sampled independently and uniformly from the unit circle in the complex plane. 

We observe two distinct phases across all architectures in Fig. \ref{fig: asymptotic 2d}: a contractive regime at low $\sigma_w$ where information fails to propagate, and a chaotic phase at higher $\sigma_w$ where information can traverse the deep network. This is similar to prior works for the fully-connected case in \cite{schoenholz2017deep, dong2022asymptotic}. Increasing the bias variance $\sigma_b$ expands the stable regime by exploiting the saturation effects of the sigmoid activation. 

\begin{figure}[t]
    \centering
    \includegraphics[width=\linewidth]{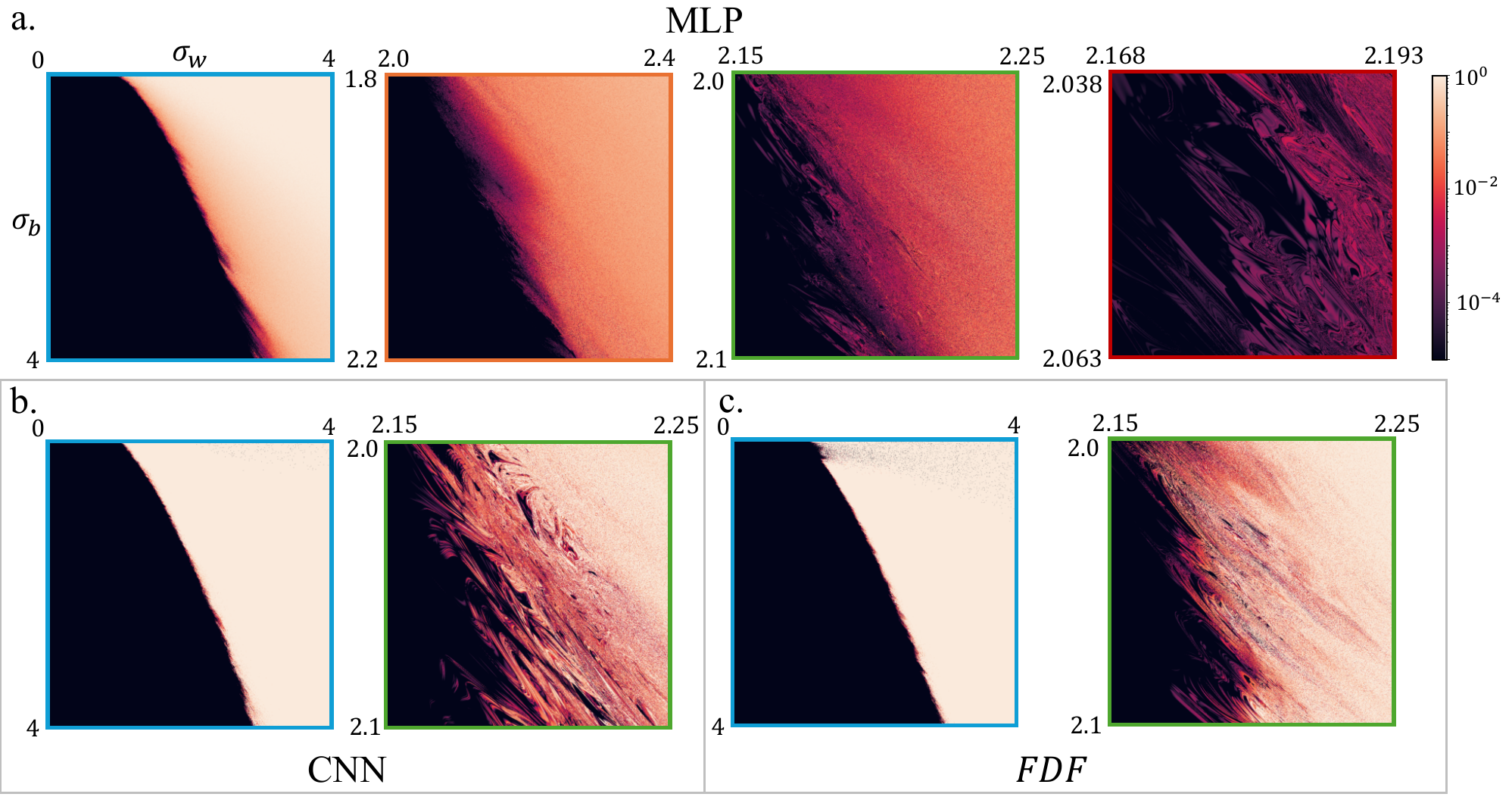}
    \caption{Zoom-in on the boundary between stability and information propagation. Sequence of 4 images of the divergence metric \old{$L^{(d)}$}\new{$L^{(D)}$} as a function of $\sigma_w$ and $\sigma_b$, with increasingly smaller range, computed for an MLP of depth $D = 10^3$ and size $N = 10^2$.}
    \label{fig:finite-size-fractals}
\end{figure}
Crucially, the boundary separating these phases remains unchanged across all architectures, apart from some finite-size effects for small $\sigma_b$. This demonstrates the universality of mean-field predictions beyond the standard i.i.d. multilayer perceptron case. Indeed, each component of $\bzd$ in Eq. \eqref{eq: mlp definition} remains Gaussian, as a linear combination of the random Gaussian vectors $\bWd$ and $\bbd$, which supports the mean-field approximation of Eq. \eqref{eq: update variance d}. Notably, a comprehensive theoretical explanation for this universality is still lacking, as existing results are largely restricted to MLPs with contractive dynamics \cite{dong2020reservoir}.

Correlations in network weights do affect finite-width networks and their convergence. As illustrated in Fig. \ref{fig:different-widths}, the information propagation landscapes for networks with widths ranging from $N = 20$ to $N = 10^3$ are visually distinct at smaller sizes between fully-connected and convolutional or structured random architectures. Nervertheless, as the width increases, these landscapes converge toward the same asymptotic frontier. 

\subsection{Fractal dimension of order-to-chaos boundary in finite-size networks}\label{sec:fractal_dimension}

\begin{figure}[ht]
    \centering
    \includegraphics[width=\linewidth]{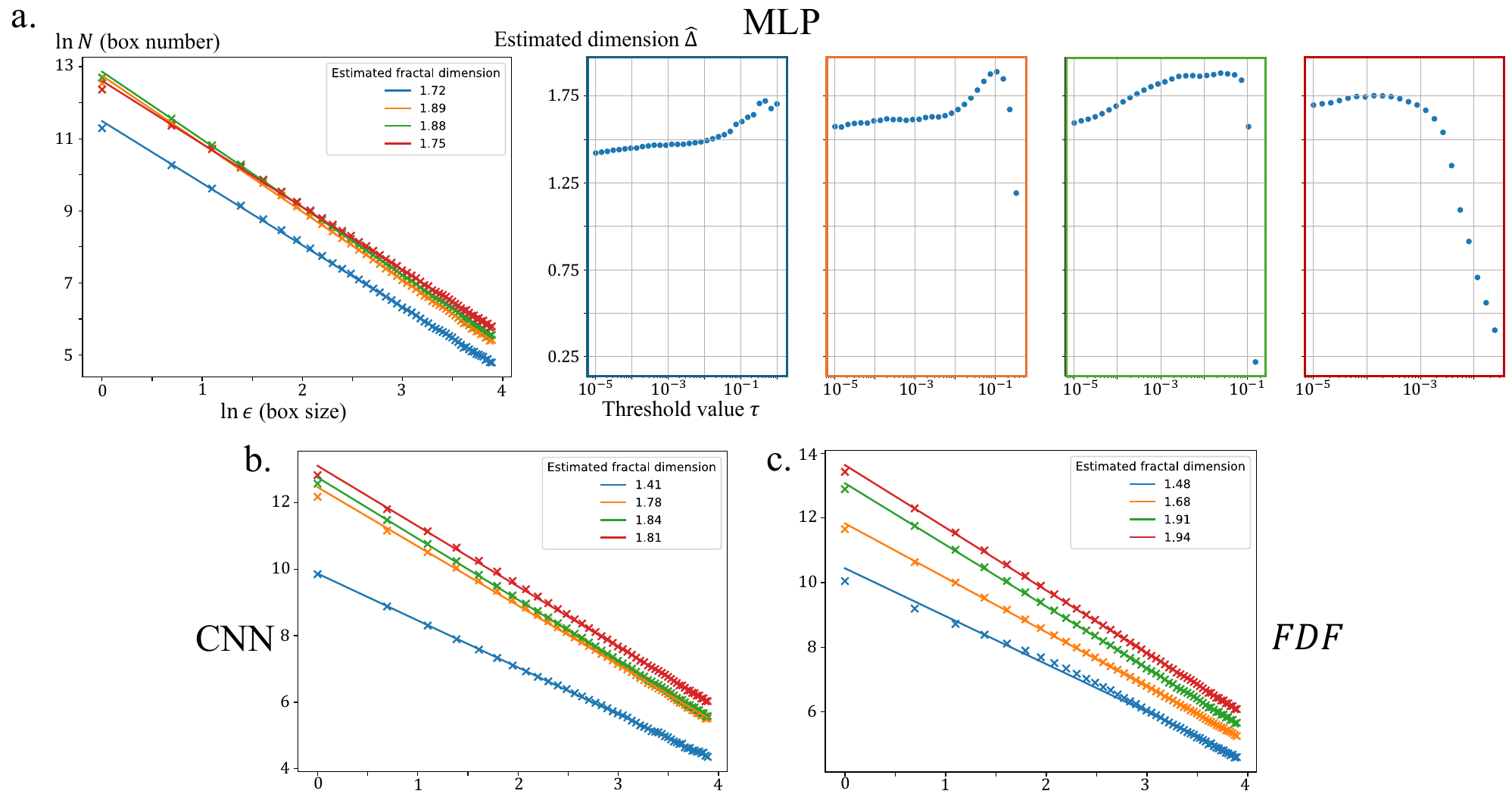}
    \caption{a) Summary of the fractal analysis of the boundary for the divergence metric \old{$L^{(d)}$}\new{$L^{(D)}$} as a function of $\sigma_w$ and $\sigma_b$, with increasingly smaller range, computed for an MLP.  Left : box size vs box counting in $\log x - \log y$ space. The estimated fractal dimension is the slope of the linear regressor, for each zooming. Right: the variation of the fractal dimension according to different thresholds $\tau$ is shown for each zoomed-in image. The threshold for the maximum fractal dimension value obtained is then chosen to carry on the analysis. The color code for each plot follows the one in Fig. \ref{fig:finite-size-fractals}, with blue corresponding to the most zoomed-out image. 
    b) Estimation of the fractal dimension for a CNN. c) Estimation of the fractal dimension for a structured random of the form $FDF$.}
    \label{fig:finite-size-fractals2}
\end{figure}
In the finite-width regime, the information propagation landscape reveals intricate fractal patterns, as illustrated in Fig. \ref{fig:finite-size-fractals}a for the MLP case. Successive zoom-ins highlight that the boundary separating ordered and chaotic phases exhibits a non-trivial structure, distinctly more complex than the smooth analytical frontier predicted by mean-field theory. Indeed, fractals often arise when iterating non-linear dynamic maps, which is consistent with the operations in a neural network. Fig. \ref{fig:finite-size-fractals}bc show that the fractal behavior is also observed as we zoom on the frontier, for the CNN and structured random architectures. 



To characterize this structure, we estimate its box dimension by first binarizing the 2D landscape with a chosen threshold $\tau$, then extracting the discrete boundary set $S_\tau$ using a gradient filter. For a range of scales $\varepsilon_j = 2^{-j}$ for $j = j_1,\dots, j_{\max}$, we count the number of boxes $N_j$ required to cover $S_\tau$, expecting a scaling law of the form:
\begin{equation}\label{eq:box_scaling}
    \ln N_j \approx C + j \hat{\Delta}. 
\end{equation}
The fractal dimension $\hat{\Delta}$ is then obtained by least-squares linear regression of $\ln N_j$ against the scale index $j$ (Figure \ref{fig:finite-size-fractals2}a, left), defined as:
\begin{equation}\label{eq:lin_reg_estimate}
    \hat{\Delta}(S_\tau) = H \quad \text{where} \quad (H,V) = \underset{(H',V')\in\mathbb{R}^2}{\arg\min} \sum_{j=j_1}^{j_{\max}} | \ln(N_j) - jH' - V'|^2.
\end{equation}

A key computational challenge is in the choice of the threshold $\tau$ defining the frontier, especially given the effects of quantization and discrete sampling. For extreme threshold values ($\tau=0$ or $\tau=1$), the boundary either covers the entire surface or is empty, while intermediate values may yield only sparse points. To address this, we define $U_\tau = \{(\sigma_r, \sigma_b): L(\sigma_r,\sigma_b) < \tau\}$ and take $S_\tau = \partial U_\tau$, identifying edges where $L$ crosses the threshold $\tau$. For a grid of threshold values $\mathcal{T} = \{\tau_1, \dots,\tau_m\}$, we compute for each one the associated fractal dimension $\hat{\Delta}(\tau)$ according to \eqref{eq:lin_reg_estimate} (Figure \ref{fig:finite-size-fractals2}a, right), and report the maximum as the fractal dimension of the frontier. 

We find a fractal dimension of approximately 1.85 for the MLP case (\ref{fig:finite-size-fractals2}a), 1.8 for the CNN case (\ref{fig:finite-size-fractals2}b), and about 1.9 for the structured random case (\ref{fig:finite-size-fractals2}c). As this value is strictly between 1 and 2, this \old{proves} \new{provides evidence of} the fractal structure of the boundary. The estimates are notably consistent for the MLP and CNN cases, whereas the structured random case exhibits a much larger spread in values. Interestingly, the estimated fractal dimension is consistently lower in the most zoomed-out images. This is likely due to the discrete sampling of $(\sigma_w, \sigma_b)$, which tends to blur finer fractal details.

\begin{figure}[t!]
    \centering
    \includegraphics[width=\linewidth]{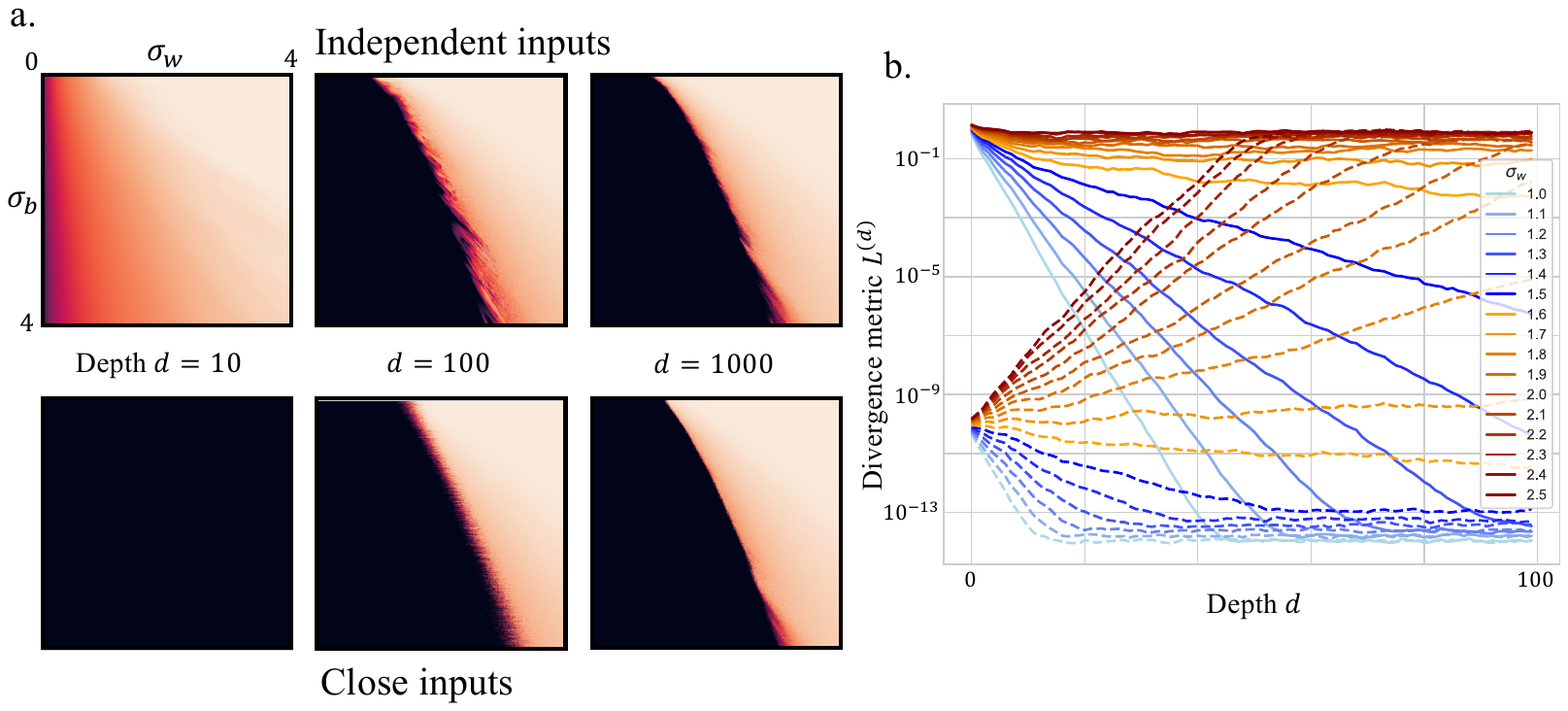}
    \caption{Comparison between of the divergence metric \new{$L^{(D)}$} for independent and close inputs. (Left) 2D images as a function of $\sigma_w$ and $\sigma_b$ for different depths. (Right) 1D line plot of the divergence metric as a functionf of depth $d$, for $\sigma_b=1$ and different values of $\sigma_w$. The independent and correlated cases correspond to the solid and dashed lines, respectively. Two regimes are observed: contractant (blue) and chaotic (brown) with an "edge of chaos" (orange) for $\sigma_w$ around 1.6.}
    \label{fig:propagation robustness tradeoff}
\end{figure}

\subsection{Separation-robustness tradeoff}\label{sec:sep_rob_tradeoff}

We extend the information propagation experiment to test network robustness by comparing the evolution of input correlations for both independent and closely related input pairs. In this setup, two inputs are considered close if one is a noise-perturbed version of the other, here with a relative difference of $10^{-5}$. This experiment evaluates the desirable property that a robust neural network should produce similar outputs for such minimally perturbed inputs, complementing the goal of information propagation, which is to effectively separate distinct inputs.

Fig. \ref{fig:propagation robustness tradeoff}a presents the divergence metric $L^{(d)}$ for both independent and close inputs across varying depths ($d=10,100,1000$). In Fig. \ref{fig:propagation robustness tradeoff}b, line plots show $L^{(d)}$ as a function of internal weight variance, with $\sigma_b = 1$ held constant. Solid lines represent the independent input case, while dashed lines correspond to close inputs.

Our results reveal that at shallow depths, the divergence metric is highly sensitive to the initial distance between inputs: close inputs remain correlated, while the distance between independent inputs is conserved. However, as the network depth increases, this distinction diminishes, and both cases converge to the same asymptotic regime characterized by two distinct phases-ordered (stable) and chaotic. This convergence explains the terminology of stable and chaotic phases: in the chaotic regime, even small input differences are amplified, leading to instability and poor noise robustness, while in the ordered regime, all inputs are mapped to similar outputs, suppressing information propagation.

Practically, this trade-off highlights the importance of finite network depth. Very deep networks fundamentally operate in two distinct regimes which are both undesirable, a chaotic phase that tends to amplify noise excessively and a stable phase that suppresses all information about inputs. Thus, initializing networks near the edge of chaos and using moderate depths provides an optimal balance between information propagation and robustness to input noise. 

\new{
\subsection{Finite-size information propagation for backpropagation}

\begin{figure}
    \centering
    \includegraphics[width=\linewidth]{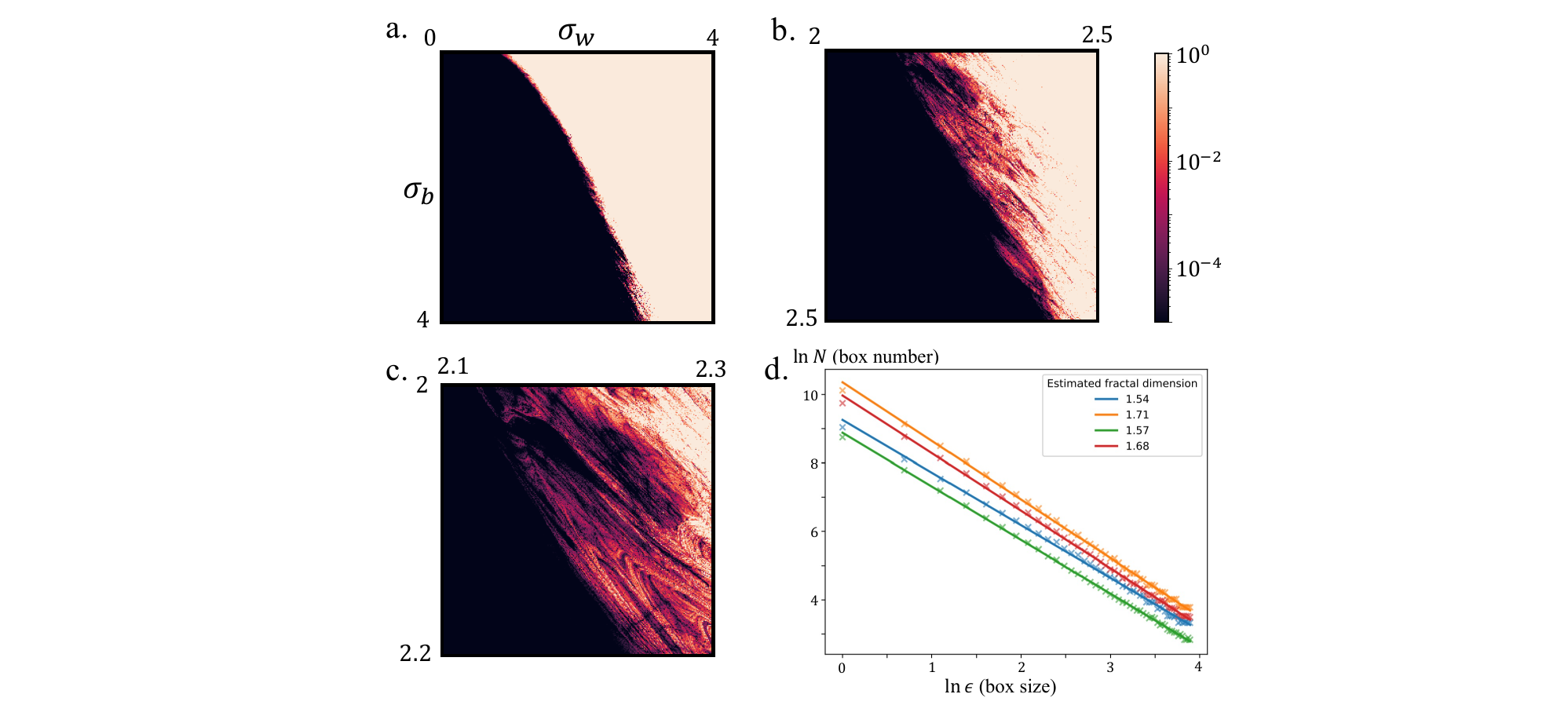}
    \caption{\new{Fractal boundary of information propagation regimes for backpropagation. a--c) Sequence of 3 images of the divergence metric $L'$ as a function of $\sigma_w$ and $\sigma_b$, computed for an MLP of depth $D=400$ and size $N=50$. d) Estimation of the fractal dimension using the box counting method, for four different zoom-ins on the fractal boundary.}}
    \label{fig:backprop}
\end{figure}

The results for backpropagation are presented in Fig. \ref{fig:backprop}. 
They have been computed for multi-layer perceptrons of width $N=50$ and depth $D=400$, with structured random connections between layers. 
Three different $300 \times 300$ zoom-ins are presented in Fig. \ref{fig:backprop}abc, the first one with $\sigma_w, \sigma_b$ both in $[0,4]$, the second in $[2,2.5]$, and the third with $\sigma_w$ in $[2.1, 2.3]$ while $\sigma_b$ is in $[2, 2.2]$.
We can observe the two distinct regimes, in which gradient information either converges to zero or propagates through the network. 

We provide in Fig. \ref{fig:backprop}.d an estimation of the box dimension of the fractal boundary, from the plots \ref{fig:backprop}.b, \ref{fig:backprop}.c and two other zoom-ins, with $\sigma_w, \sigma_b$ both in $[1.5, 2.5]$ for one, and $\sigma_w$ in $[2.25, 2.45]$ and $\sigma_b$ in $[2.3, 2.5]$ for the other. 
The thresholds have been chosen according to the method described in Fig. \ref{fig:finite-size-fractals2}. 
Interestingly, they present the same transition between stability and propagation regimes, with a fractal boundary of dimension around 1.6. 
This demonstrates the complexity of the dynamics of the propagation of information also in the backward pass. 
}

\section{Discussion and perspectives}\label{sec:discussion_perspectives}

In this work, we have demonstrated the emergence of fractal behavior in deep information propagation within finite-width neural networks. By systematically investigating architectures ranging from MLPs to CNNs and analyzing the effects of internal weight and bias variances, we estimated the fractal dimension of the boundary separating stable and chaotic regimes in the finite-size context. Our study also explored the tradeoff between separation and robustness, emphasizing the critical role of network depth. \new{We also displayed that similar fractal boundaries are observed when looking at information backpropagation. The source code is freely available as a public repository\footnote{\url{https://github.com/jon-dong/fractal-deep-info-prop}}.}

Looking ahead, our findings open several promising directions for future research. One avenue is the theoretical characterization of the mean-field approximation beyond the contractant MLP case. The universality observed in practice suggest that convergence can be rigorously proven with far fewer assumptions. \new{Such characterization may be addressed for instance exploiting the concept of Dynamic Mean Field (DMF) \cite{crisanti2018path,bordelon2023dynamics}.} Moreover, theoretical studies through the lens of dynamical systems theory could shed light on information propagation in finite-size neural networks. 

Additionally, extending our analysis to widely used architectural and training heuristics---such as Xavier initialization, residual connections, dropout, and layer normalization---could yield valuable practical insights. Finally, applying our framework to more complex and contemporary architectures, including Transformers \cite{vaswani2017attention} and Graph Neural Networks \cite{scarselli2008graph}, represents an exciting opportunity to further advance our understanding of information dynamics in modern deep learning systems.

\section*{Acknowledgements}
G.A.D. and G.R. acknowledge the support provided by the European Union -- NextGenerationEU, in the framework of the iNEST -- Interconnected Nord-Est Innovation Ecosystem (iNEST ECS00000043 – CUP G93C22000610007) project and its CC5 Young Researchers initiative. G.R. acknowledge the support provided by PRIN "FaReX - Full and Reduced order modelling of coupled systems: focus on non-matching methods and automatic learning" project, and by INdAM-GNCS 2019–2020 projects and PON "Research and Innovation on Green related issues" FSE REACT-EU 2021 project. 
Z.H. and J.D. acknowledge funding funding from the Swiss National Science Foundation (Grant PZ00P2\_216211).
The views and opinions expressed are solely those of the authors and do not necessarily reflect those of the European Union, nor can the European Union be held responsible for them. G.A.D. and G.R. would like to acknowledge INdAM–GNCS.

 \bibliographystyle{model1-num-names} 
 \bibliography{references}

\begin{thebibliography}{22}
\expandafter\ifx\csname natexlab\endcsname\relax\def\natexlab#1{#1}\fi
\providecommand{\url}[1]{\texttt{#1}}
\providecommand{\href}[2]{#2}
\providecommand{\path}[1]{#1}
\providecommand{\DOIprefix}{doi:}
\providecommand{\ArXivprefix}{arXiv:}
\providecommand{\URLprefix}{URL: }
\providecommand{\Pubmedprefix}{pmid:}
\providecommand{\doi}[1]{\href{http://dx.doi.org/#1}{\path{#1}}}
\providecommand{\Pubmed}[1]{\href{pmid:#1}{\path{#1}}}
\providecommand{\bibinfo}[2]{#2}
\ifx\xfnm\relax \def\xfnm[#1]{\unskip,\space#1}\fi
\bibitem[{Schoenholz et~al.(2017)Schoenholz, Gilmer, Ganguli, and
  Sohl-Dickstein}]{schoenholz2017deep}
\bibinfo{author}{S.~S. Schoenholz}, \bibinfo{author}{J.~Gilmer},
  \bibinfo{author}{S.~Ganguli}, \bibinfo{author}{J.~Sohl-Dickstein},
\newblock \bibinfo{title}{Deep information propagation},
\newblock in: \bibinfo{booktitle}{International Conference on Learning
  Representations}, \bibinfo{year}{2017}.
\bibitem[{Poole et~al.(2016)Poole, Lahiri, Raghu, Sohl-Dickstein, and
  Ganguli}]{poole2016exponential}
\bibinfo{author}{B.~Poole}, \bibinfo{author}{S.~Lahiri},
  \bibinfo{author}{M.~Raghu}, \bibinfo{author}{J.~Sohl-Dickstein},
  \bibinfo{author}{S.~Ganguli},
\newblock \bibinfo{title}{Exponential expressivity in deep neural networks
  through transient chaos},
\newblock \bibinfo{journal}{Advances in Neural Information Processing systems}
  \bibinfo{volume}{29} (\bibinfo{year}{2016}).
\bibitem[{Dong et~al.(2020)Dong, Ohana, Rafayelyan, and
  Krzakala}]{dong2020reservoir}
\bibinfo{author}{J.~Dong}, \bibinfo{author}{R.~Ohana},
  \bibinfo{author}{M.~Rafayelyan}, \bibinfo{author}{F.~Krzakala},
\newblock \bibinfo{title}{Reservoir computing meets recurrent kernels and
  structured transforms},
\newblock \bibinfo{journal}{Advances in Neural Information Processing Systems}
  \bibinfo{volume}{33} (\bibinfo{year}{2020}) \bibinfo{pages}{16785--16796}.
\bibitem[{Dong et~al.(2022)Dong, B{\"o}rve, Rafayelyan, and
  Unser}]{dong2022asymptotic}
\bibinfo{author}{J.~Dong}, \bibinfo{author}{E.~B{\"o}rve},
  \bibinfo{author}{M.~Rafayelyan}, \bibinfo{author}{M.~Unser},
\newblock \bibinfo{title}{Asymptotic stability in reservoir computing},
\newblock in: \bibinfo{booktitle}{2022 International Joint Conference on Neural
  Networks (IJCNN)}, \bibinfo{organization}{IEEE}, \bibinfo{year}{2022}, pp.
  \bibinfo{pages}{01--08}.
\bibitem[{D’Inverno and Dong(2025)}]{d2025comparison}
\bibinfo{author}{G.~A. D’Inverno}, \bibinfo{author}{J.~Dong},
\newblock \bibinfo{title}{Comparison of reservoir computing topologies using
  the recurrent kernel approach},
\newblock \bibinfo{journal}{Neurocomputing} \bibinfo{volume}{611}
  (\bibinfo{year}{2025}) \bibinfo{pages}{128679}.
\bibitem[{Sohl-Dickstein(2024)}]{sohl2024boundary}
\bibinfo{author}{J.~Sohl-Dickstein},
\newblock \bibinfo{title}{The boundary of neural network trainability is
  fractal},
\newblock \bibinfo{journal}{arXiv preprint arXiv:2402.06184}
  (\bibinfo{year}{2024}).
\bibitem[{Torkamandi(2025)}]{torkamandi2025mapping}
\bibinfo{author}{B.~Torkamandi},
\newblock \bibinfo{title}{Mapping the edge of chaos: Fractal-like boundaries in
  the trainability of decoder-only transformer models},
\newblock \bibinfo{journal}{arXiv preprint arXiv:2501.04286}
  (\bibinfo{year}{2025}).
\bibitem[{Hu et~al.(2025)Hu, Tachella, Unser, and Dong}]{hu2025structured}
\bibinfo{author}{Z.~Hu}, \bibinfo{author}{J.~Tachella},
  \bibinfo{author}{M.~Unser}, \bibinfo{author}{J.~Dong},
\newblock \bibinfo{title}{Structured random model for fast and robust phase
  retrieval},
\newblock in: \bibinfo{booktitle}{ICASSP 2025-2025 IEEE International
  Conference on Acoustics, Speech and Signal Processing (ICASSP)},
  \bibinfo{organization}{IEEE}, \bibinfo{year}{2025}, pp.
  \bibinfo{pages}{1--5}.
\bibitem[{Yu et~al.(2016)Yu, Suresh, Choromanski, Holtmann-Rice, and
  Kumar}]{yu2016orthogonal}
\bibinfo{author}{F.~X.~X. Yu}, \bibinfo{author}{A.~T. Suresh},
  \bibinfo{author}{K.~M. Choromanski}, \bibinfo{author}{D.~N. Holtmann-Rice},
  \bibinfo{author}{S.~Kumar},
\newblock \bibinfo{title}{Orthogonal random features},
\newblock \bibinfo{journal}{Advances in Neural Information Processing Systems}
  \bibinfo{volume}{29} (\bibinfo{year}{2016}).
\bibitem[{Jaeger(2001)}]{jaeger2001echo}
\bibinfo{author}{H.~Jaeger},
\newblock \bibinfo{title}{The “echo state” approach to analysing and
  training recurrent neural networks-with an erratum note},
\newblock \bibinfo{journal}{Bonn, Germany: German National Research Center for
  Information Technology GMD Technical Report} \bibinfo{volume}{148}
  (\bibinfo{year}{2001}) \bibinfo{pages}{13}.
\bibitem[{Luko{\v{s}}evi{\v{c}}ius et~al.(2012)Luko{\v{s}}evi{\v{c}}ius,
  Jaeger, and Schrauwen}]{lukovsevivcius2012reservoir}
\bibinfo{author}{M.~Luko{\v{s}}evi{\v{c}}ius}, \bibinfo{author}{H.~Jaeger},
  \bibinfo{author}{B.~Schrauwen},
\newblock \bibinfo{title}{Reservoir computing trends},
\newblock \bibinfo{journal}{KI-K{\"u}nstliche Intelligenz} \bibinfo{volume}{26}
  (\bibinfo{year}{2012}) \bibinfo{pages}{365--371}.
\bibitem[{Tanaka et~al.(2019)Tanaka, Yamane, H{\'e}roux, Nakane, Kanazawa,
  Takeda, Numata, Nakano, and Hirose}]{tanaka2019recent}
\bibinfo{author}{G.~Tanaka}, \bibinfo{author}{T.~Yamane},
  \bibinfo{author}{J.~B. H{\'e}roux}, \bibinfo{author}{R.~Nakane},
  \bibinfo{author}{N.~Kanazawa}, \bibinfo{author}{S.~Takeda},
  \bibinfo{author}{H.~Numata}, \bibinfo{author}{D.~Nakano},
  \bibinfo{author}{A.~Hirose},
\newblock \bibinfo{title}{Recent advances in physical reservoir computing: A
  review},
\newblock \bibinfo{journal}{Neural Networks} \bibinfo{volume}{115}
  (\bibinfo{year}{2019}) \bibinfo{pages}{100--123}.
\bibitem[{Rafayelyan et~al.(2020)Rafayelyan, Dong, Tan, Krzakala, and
  Gigan}]{rafayelyan2020large}
\bibinfo{author}{M.~Rafayelyan}, \bibinfo{author}{J.~Dong},
  \bibinfo{author}{Y.~Tan}, \bibinfo{author}{F.~Krzakala},
  \bibinfo{author}{S.~Gigan},
\newblock \bibinfo{title}{Large-scale optical reservoir computing for
  spatiotemporal chaotic systems prediction},
\newblock \bibinfo{journal}{Physical Review X} \bibinfo{volume}{10}
  (\bibinfo{year}{2020}) \bibinfo{pages}{041037}.
\bibitem[{Moczulski et~al.(2015)Moczulski, Denil, Appleyard, and
  de~Freitas}]{moczulski2015acdc}
\bibinfo{author}{M.~Moczulski}, \bibinfo{author}{M.~Denil},
  \bibinfo{author}{J.~Appleyard}, \bibinfo{author}{N.~de~Freitas},
\newblock \bibinfo{title}{Acdc: A structured efficient linear layer},
\newblock \bibinfo{journal}{arXiv preprint arXiv:1511.05946}
  (\bibinfo{year}{2015}).
\bibitem[{Dao et~al.(2019)Dao, Gu, Eichhorn, Rudra, and
  R{\'e}}]{dao2019learning}
\bibinfo{author}{T.~Dao}, \bibinfo{author}{A.~Gu},
  \bibinfo{author}{M.~Eichhorn}, \bibinfo{author}{A.~Rudra},
  \bibinfo{author}{C.~R{\'e}},
\newblock \bibinfo{title}{Learning fast algorithms for linear transforms using
  butterfly factorizations},
\newblock in: \bibinfo{booktitle}{International Conference on Machine
  Learning}, \bibinfo{organization}{PMLR}, \bibinfo{year}{2019}, pp.
  \bibinfo{pages}{1517--1527}.
\bibitem[{Falconer(2006)}]{falconer_fractal_2006}
\bibinfo{author}{K.~J. Falconer}, \bibinfo{title}{Fractal geometry:
  mathematical foundations and applications}, \bibinfo{edition}{2. ed., repr}
  ed., \bibinfo{publisher}{Wiley}, \bibinfo{address}{Chichester},
  \bibinfo{year}{2006}.
\bibitem[{Strogatz(2018)}]{strogatz2018nonlinear}
\bibinfo{author}{S.~H. Strogatz}, \bibinfo{title}{Nonlinear dynamics and chaos:
  with applications to physics, biology, chemistry, and engineering},
  \bibinfo{publisher}{CRC press}, \bibinfo{year}{2018}.
\bibitem[{Abry et~al.(2009)Abry, Gonçalves, and
  Lévy~Véhel}]{abry2009scaling}
\bibinfo{editor}{P.~Abry}, \bibinfo{editor}{P.~Gonçalves},
  \bibinfo{editor}{J.~Lévy~Véhel} (Eds.), \bibinfo{title}{Scaling, fractals
  and wavelets}, Digital signal and image processing series,
  \bibinfo{publisher}{Wiley-ISTE}, \bibinfo{address}{London},
  \bibinfo{year}{2009}.
\bibitem[{Crisanti and Sompolinsky(2018)}]{crisanti2018path}
\bibinfo{author}{A.~Crisanti}, \bibinfo{author}{H.~Sompolinsky},
\newblock \bibinfo{title}{Path integral approach to random neural networks},
\newblock \bibinfo{journal}{arXiv preprint arXiv:1809.06042}
  (\bibinfo{year}{2018}).
\bibitem[{Bordelon and Pehlevan(2023)}]{bordelon2023dynamics}
\bibinfo{author}{B.~Bordelon}, \bibinfo{author}{C.~Pehlevan},
\newblock \bibinfo{title}{Dynamics of finite width kernel and prediction
  fluctuations in mean field neural networks},
\newblock \bibinfo{journal}{Advances in Neural Information Processing Systems}
  \bibinfo{volume}{36} (\bibinfo{year}{2023}) \bibinfo{pages}{9707--9750}.
\bibitem[{Vaswani et~al.(2017)Vaswani, Shazeer, Parmar, Uszkoreit, Jones,
  Gomez, Kaiser, and Polosukhin}]{vaswani2017attention}
\bibinfo{author}{A.~Vaswani}, \bibinfo{author}{N.~Shazeer},
  \bibinfo{author}{N.~Parmar}, \bibinfo{author}{J.~Uszkoreit},
  \bibinfo{author}{L.~Jones}, \bibinfo{author}{A.~N. Gomez},
  \bibinfo{author}{{\L}.~Kaiser}, \bibinfo{author}{I.~Polosukhin},
\newblock \bibinfo{title}{Attention is all you need},
\newblock \bibinfo{journal}{Advances in Neural Information Processing Systems}
  \bibinfo{volume}{30} (\bibinfo{year}{2017}).
\bibitem[{Scarselli et~al.(2008)Scarselli, Gori, Tsoi, Hagenbuchner, and
  Monfardini}]{scarselli2008graph}
\bibinfo{author}{F.~Scarselli}, \bibinfo{author}{M.~Gori},
  \bibinfo{author}{A.~C. Tsoi}, \bibinfo{author}{M.~Hagenbuchner},
  \bibinfo{author}{G.~Monfardini},
\newblock \bibinfo{title}{{The Graph Neural Network Model}},
\newblock \bibinfo{journal}{IEEE Transactions on Neural Networks}
  \bibinfo{volume}{20} (\bibinfo{year}{2008}) \bibinfo{pages}{61--80}.

\end{thebibliography}

\end{document}